\documentclass{article}

\newcommand{\by}{\boldsymbol{y}}
\newcommand{\bz}{\boldsymbol{z}}

\newcommand{\bT}{\boldsymbol{T}}

\newcommand{\bq}{\boldsymbol{q}}
\newcommand{\bk}{\boldsymbol{k}}
\newcommand{\bK}{\boldsymbol{K}}

\newcommand{\ba}{\boldsymbol{a}}

\newcommand{\bV}{\boldsymbol{V}}
\newcommand{\bP}{\boldsymbol{P}}

\newcommand{\bC}{\boldsymbol{C}}

\usepackage{graphicx} 
\usepackage{subcaption}
\usepackage{algorithm}
\usepackage{algpseudocode}
\usepackage{amsmath}
\usepackage{booktabs} % 用于专业表格线型
\usepackage{array}    % 用于列格式调整
\usepackage{multirow} % 合并单元格（未使用但常用）
\usepackage{tabularx}
\usepackage[table,x11names]{xcolor}
\definecolor{myGreen}{RGB}{213, 232, 212}
\definecolor{myCyan}{RGB}{218, 232, 252}

\usepackage[style=numeric,maxnames=999,giveninits=true, backend=biber]{biblatex}
\usepackage{tikz}
\usepackage{amsmath}
\usetikzlibrary{arrows.meta, positioning}
\usepackage[preprint]{neurips_2025}

\usepackage[utf8]{inputenc} % allow utf-8 input
\usepackage[T1]{fontenc}    % use 8-bit T1 fonts
\usepackage{hyperref}       % hyperlinks
\usepackage{url}            % simple URL typesetting
\usepackage{booktabs}       % professional-quality tables
\usepackage{amsfonts}       % blackboard math symbols
\usepackage{nicefrac}       % compact symbols for 1/2, etc.
\usepackage{microtype}      % microtypography
\usepackage{xcolor}         % colors
\usepackage[inline]{enumitem}

\newcommand{\ours}{HCAttention}

\title{HCAttention: An Extreme KV Cache Compression for Long-Context LLMs: A Heterogeneous Attention Computing with Key Quantization and Value Offloading}

\title{HCAttention: An Extreme KV Cache Compression for Long-Context LLMs via A Heterogeneous Attention Computing}

\title{HCAttention: Extreme KV Cache Compression via Heterogeneous Attention Computing for LLMs}

\author{%
  Dongquan Yang
    \\
  Intellifusion Inc.\\
  Shenzhen, China \\
  \texttt{matrixyangdq@gmail.com} \\
  \And
  Yifan Yang \\
  Intellifusion Inc. \\
  Shenzhen, China \\
  \texttt{yifan.yang.cn@gmail.com} \\
  \AND
  Xiaotian Yu \\
  Intellifusion Inc. \\
  Shenzhen, China \\
  \texttt{xiaotianyu.ac@gmail.com} \\
  \And
  Xianbiao Qi \\
  Intellifusion Inc. \\
  Shenzhen, China \\
  \texttt{qixianbiao@gmail.com} \\
  \And
  Rong Xiao \\
  Intellifusion Inc. \\
  Shenzhen, China \\
  \texttt{rongxiao@gmail.com} \\
}

\begin{document}

\maketitle

\begin{abstract}
Processing long-context inputs with large language models presents a significant challenge due to the enormous memory requirements of the Key-Value (KV) cache during inference. Existing KV cache compression methods exhibit noticeable performance degradation when memory is reduced by more than 85\%. Additionally, strategies that leverage GPU-CPU collaboration for approximate attention remain underexplored in this setting.
We propose \ours, a heterogeneous attention computation framework that integrates key quantization, value offloading, and dynamic KV eviction to enable efficient inference under extreme memory constraints. The method is compatible with existing transformer architectures and does not require model fine-tuning.
Experimental results on the LongBench benchmark demonstrate that our approach preserves the accuracy of full-attention model while shrinking the KV cache memory footprint to 25\% of its original size. Remarkably, it stays competitive with only 12.5\% of the cache, setting a new state-of-the-art in LLM KV cache compression. To the best of our knowledge, \ours~is the first to extend the Llama-3-8B model to process 4 million tokens on a single A100 GPU with 80GB memory.

\end{abstract}
\section{Introduction}
% \paragraph{LLM in long contexts}
Large Language Models (LLMs)~\cite{achiam2023gpt,bai2023qwen,liu2024deepseek,touvron2023llama,ni2021sentence,grattafiori2024llama,team2023gemini,jiang2024mixtral} have demonstrated remarkable capabilities across diverse natural language processing tasks, but deploying them on long contexts poses serious efficiency challenges. Many real-world applications, including multi-turn dialogue~\cite{zhang2025survey,li2025beyond}, document understanding~\cite{li2024focusllm,zhang2024soaring}, and AI agent~\cite{zhang2024chain,mao2024lift}, necessitate extending the context lengths of LLMs. The memory overhead of self-attention grows linearly with sequence length because the model must store a key and value vector for every past token. This quickly becomes prohibitive. For example, the Key-Value (KV) cache for the OPT-175B model~\cite{zhang2022opt}, with 175 billion parameters, can demand approximately 950GB of memory when processing a batch size of 128 and a sequence length of 2,048 – roughly three times the size of the model's parameters.
Even with memory swapping and optimized libraries, inference on long sequences can become memory-bound, hurting throughput and latency. There is a pressing need for techniques that drastically reduce KV cache memory usage without sacrificing model performance.
%This is primarily due to their ability to effectively capture and utilize long-range dependencies through the self-attention mechanism. Many real-world applications, including multi-turn dialogue~\cite{zhang2025survey,li2025beyond}, document understanding~\cite{li2024focusllm,zhang2024soaring}, and AI agent~\cite{zhang2024chain,mao2024lift}, necessitate extending the context lengths of LLMs. However, deploying LLMs in these long-context scenarios, potentially involving sequences extending to millions of tokens, faces significant efficiency barriers.

The challenges stem from the inherent complexity of the self-attention mechanism~\cite{vaswani2017attention}, which incurs substantial costs along two primary dimensions: computational complexity, which scales quadratically with sequence length, and memory complexity, specifically related to the KV cache. The memory size of KV cache expands linearly with the context length, leading to a substantial increase in memory consumption. This substantial memory footprint not only limits the maximum processable sequence length and achievable batch size but also impedes efficient concurrent deployment. %Consequently, reducing KV cache memory consumption is a critical requirement for achieving efficient, scalable, and widely accessible LLM inference.

% Transformer-based LLMs necessitate caching key and value (KV) embeddings of all preceding tokens to facilitate attention over extended sequences. While crucial for comprehensive contextual reasoning, this design exhibits unfavorable scaling characteristics in both computational time and memory footprint. With the ongoing trend towards longer context windows, the resources dedicated to storing and retrieving KV caches increasingly dictate overall inference costs. 
% % For instance, a 7-billion parameter model with 32 layers processing a 32k-token context can accumulate a KV cache exceeding 20GB. 
% This substantial memory overhead not only restricts the maximum processable sequence length and achievable batch size but also hinders the concurrent deployment. Consequently, mitigating KV cache memory consumption is crucial for efficient, scalable, and broadly accessible LLM inference.

% \textcolor{red}{TODO: add related work,  motivations and observations}
Prior research has significantly advanced the efficiency of LLM inference, especially in handling long input sequences~\cite{vyas2020fast, ge2023model,liu2024deepseek,xiao2024duoattention}. 
Findings from ~\cite{adafactor_shazeer2018adafactor} suggest that not all tokens are equally critical for attention computation. Building upon this insight, recent methods estimate token importance using attention scores or derived statistical heuristics~\cite{xiao2024duoattention, wang2025squeezeattention, xu2024think}. For example, Longformer~\cite{beltagy2020longformer} introduced sliding window attention combined with global tokens to enable efficient LLM applications, while DuoAttention~\cite{xiao2024duoattention} applied the eviction strategy to only a subset of attention heads for managing long-sequence generation.  
However, these methods suffer from significant performance degradation when attempting to further reduce the KV cache footprint beyond 85\%. Even worse, the permanent-eviction strategies in prior studies introduce a systematic flaw. That is, tokens discarded early may later prove essential, resulting in substantial accuracy drops.

Pushing the boundary of KV cache compression ratio requires addressing a series of interrelated challenges. A central problem is how to dynamically evict redundant KV entries while preserving all tokens essential for maintaining model performance throughout the entire inference process. Unlike static truncation (or token selection), effective token eviction must adapt to changing token importance over time and across layers, which is difficult to assess in an online setting.

In order to achieve dynamic token eviction, a potential solution is to take advantage of cache quantization, e.g., KIVI \cite{liu2024kivi} and Coupled Quantization \cite{zhang2024kv}. Besides, among all KV components, it is noticeable that value vectors dominate the memory budget even though they are accessed only during the final weighted-sum step. This makes them prime candidates for CPU offloading. However, the heterogeneous computing between CPUs and GPUs in KV cache compression of LLMs is underexplored. These observations highlight a key open question: how to design efficient value offloading and approximate attention mechanisms with extreme KV cache compression without sacrificing inference speed or model accuracy.

To address the aforementioned challenges, we argue that a unified framework must holistically coordinate multiple complementary techniques. Specifically, we identify three key strategies that, when integrated, can effectively overcome the limitations of existing KV cache compression approaches. 
First, key quantization can significantly reduce the memory footprint of the KV cache by compressing high-dimensional key vectors. This tackles the performance-memory trade-off while preserving all essential tokens during inference. Second, value offloading shifts memory-intensive value vectors to CPU memory, leveraging the CPU’s abundant capacity. Third, KV eviction selectively discards low-contribution tokens in real time, maintaining the integrity of necessary computations.

To the best of our knowledge, this is the first unified framework that simultaneously leverages key quantization, value offloading, and dynamic KV eviction within a heterogeneous GPU-CPU architecture for efficient long-context inference. Prior works have explored these techniques in isolation, but none have effectively integrated them into a cohesive system that balances memory efficiency, computational cost, and inference quality. We posit that only a cohesive, device-aware design that leverages both GPUs and CPUs can push the limits of KV cache compression without sacrificing model performance.

Based on this insight, we propose \ours, a heterogeneous approximate attention computation framework that fully exploits the complementary strengths of GPU and CPU. \ours~extends the Llama-3-8B model to process 4 million tokens on a single A100 GPU with 80GB memory.  In our framework, GPUs perform high-throughput attention score computation over quantized keys, while the CPU absorbs offloaded values and participates in final computation. Dynamic eviction policies further reduce memory load by pruning non-essential KV pairs across layers based on cumulative magnitude criterion. Our contributions are three-fold:
\setlist{leftmargin=*}
\begin{itemize}
\item We propose an approximate attention mechanism that combines key quantization and dynamic KV eviction, enabling high compression efficiency with minimal accuracy loss.
\item We introduce a heterogeneous GPU-CPU execution architecture that offloads value cache storage and computation to CPU memory via overlap-aware, asynchronous coordination, ensuring runtime efficiency.
\item We demonstrate through experiments on long-context benchmarks that \ours~can reduce the KV cache memory footprint to 25\% of the original size while maintaining the performance of full-attention model. Even with only 12.5\% of the original cache, our method remains highly competitive, achieving less than 1\% degradation in accuracy.
\end{itemize}

% Furthermore, our proposed method exhibits compatibility with recent attention optimizations, such as Multi-head Latent Attention (MLA) as implemented in models like DeepSeek. When integrated with MLA, our approach achieves a substantial reduction in total KV cache memory footprint, retaining only 3.5\% of the original memory.

\section{Related Work}
\label{related_work}
 Efficient KV cache management for scalable LLM deployment has been explored in the following three categories: KV caches selection, KV caches quantization and heterogeneous computing of LLM inference.

\paragraph{KV caches selection}
Recent studies focus on identifying and retaining only important tokens to reduce the size of KV caches.
TOVA~\cite{oren2024transformers} proposes Token Omission Via Attention, which at each step drops the token with the lowest attention score to the current query, effectively keeping only the most attended values in the state.
StreamingLLM~\cite{xiao2023efficient} targets streaming inference by combining a sliding window of recent tokens with an “attention sink” mechanism. It retains certain early tokens (which tend to attract strong long-range attention) as anchors, allowing an effectively unbounded context without fine-tuning.
Heavy--Hitter Oracle (H2O)~\cite{zhang2023h2o} observes that a small subset of ``heavy‐hitter'' tokens, those with high cumulative attention, dominate the value computation, and it dynamically select but permanently evicts cache entries to keep a balance of these heavy hitters and the most recent tokens. 
Keyformer~\cite{adnan2024keyformer} introduces a learned token selection mechanism based on Gumbel-softmax sampling. Keyformer scores tokens according to their contextual importance and selectively keeps a subset of keys and values. The permanent eviction of caches in previous methods might incur accuracy drops.
Later, SqueezeAttention~\cite{wang2025squeezeattention} further proposes a hierarchical memory budget allocation across transformer layers, coupled with adaptive sequence pruning algorithms per layer. 
Similarly, DuoAttention~\cite{xiao2024duoattention} further refines this by splitting attention heads into ``retrieval'' heads (given full KV cache) versus ``streaming'' heads (given a small fixed cache), based on an optimization that identifies which heads need long range memory.
The Quest~\cite{tang2024quest} method takes a query aware approach. It precomputes min/max summaries of each KV-page and, for each new query, estimates a criticality score for pages via inner products.

\paragraph{KV caches quantization}
Latest work has proposed a variety of ultra-low-bit schemes for KV-cache quantization.
KIVI~\cite{liu2024kivi} characterizes the statistics of KV caches and finds that keys should be quantized per-channel whereas values per-token. Similarly, AsymKV~\cite{tao2025asymkv} also considers the asymmetric configuration of quantization between key and value matrices at the layer level.
ZipCache~\cite{he2024zipcache} proposes to represent important tokens receive fewer bits, using normalized attention-score metric to evaluate token saliency.  Coupled Quantization~\cite{zhang2024kv} observes that channels of key and value exhibit high level of dependency. This method scheme jointly quantize groups of channels rather than treating each independently, encoding them more information-efficiently. However, none of these approaches systematically investigates how token selection policies could further amplify the benefits of quantization.

% By assigning distinct bit allocations layerwise (typically more bits for keys, fewer for values).

\paragraph{Heterogeneous computing}
Recent heterogeneous computing approaches for KV cache demonstrate diverse strategies for optimizing CPU-GPU collaboration. For instance, systems like NEO~\cite{jiang2024neo} and FastDecode~\cite{he2024fastdecode} implement strategic workload distribution through CPU offloading of attention computations. In contrast, FlexInfer~\cite{xu2024vtensor} introduces virtual memory abstractions to enhance resource coordination.
While these methods primarily emphasize load-aware scheduling and asymmetric pipelining, our approach centers on minimizing both communication and computation overhead to effectively utilize CPUs for KV cache computations and value cache storage.

\textbf{Remark.} \ours~lies in the intersection of the above three directions. It targets at the KV-cache optimization. We use a quantization strategy to quant the keys. For the values, we offload them to the CPU so it will largely save GPU memory. To compute attention, we uses a heterogeneous computing strategy with part of the attention equation is calculated on GPU and the other part is computed on CPU.

\section{\ours}
\label{sec:method}
\begin{figure}[t]
    \centering
    \includegraphics[width=1\linewidth]{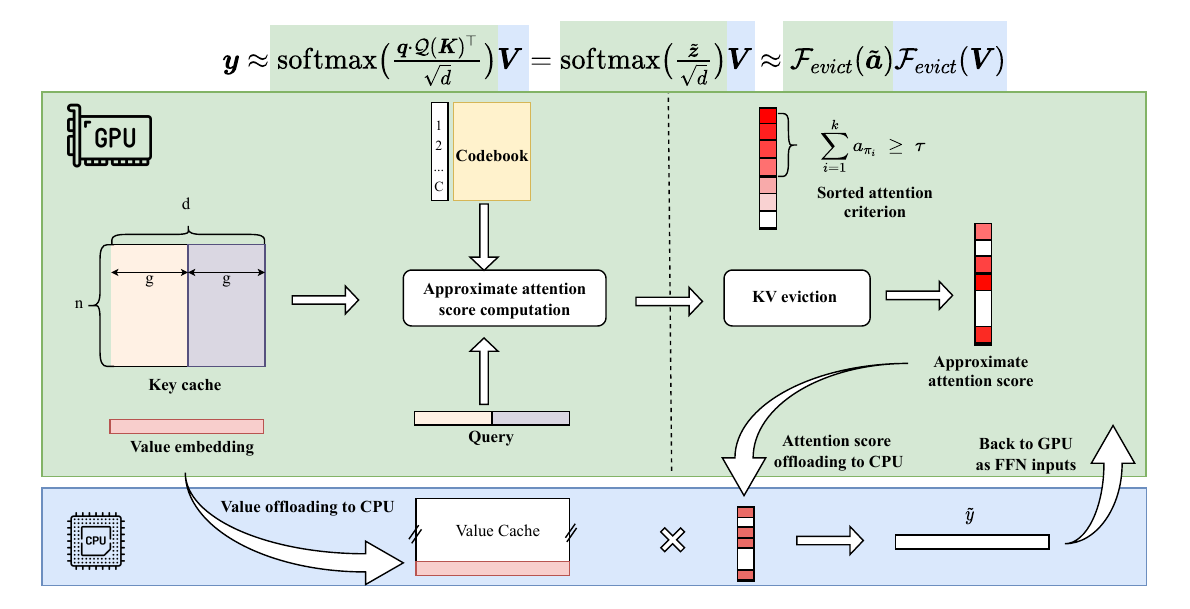}
    \caption{Overview of the \ours~heterogeneous attention computation framework. The unified system comprises three key components: (1) approximate attention score computation via key quantization on GPU, (2) KV eviction based on a cumulative magnitude criterion, and (3) final context vector computation with value offloading to CPU. Computational and storage resources represented in two colors: green indicates GPU, and light blue indicates CPU.}
    \label{fig:overview}
\end{figure}

\ours~introduces a novel heterogeneous attention computation framework that enables extreme KV cache compression with minimal impact on inference quality.
% Despite aggressive compression, it maintains inference performance through carefully coordinated resource usage and approximation strategies.
As illustrated in Fig.~\ref{fig:overview}, our holistic strategy comprises three key components:
(1) quantization-driven approximate attention computation;
% : We apply key cache quantization to enable memory-efficient attention computation. This reduces the GPU memory footprint while preserving attention accuracy, facilitating faster computation with approximate representations.
(2) dynamic KV eviction based on a cumulative magnitude criterion;
% : To minimize computation overhead on CPU, we selectively transfer the most salient tokens between GPU and CPU. This significantly reduces bandwidth usage and computational cost, ensuring efficient offloading without stalling the computation pipeline.
(3) heterogeneous attention execution with value cache offloading.
% : To further relieve GPU memory pressure, we offload the value cache and parts of the attention computation to CPU memory. This design leverages the CPU's ample memory capacity and the GPU's computational efficiency, enabling seamless collaboration across devices.

%To show our method
\subsection{Key Quantization}
\label{sec:key_quantization}
Inspired by Zhang et al.~\cite{zhang2024kv}, we utilize grouped vector quantization for the key cache, which takes full advantage of interdependencies among key embedding dimensions. The quantization codebook is constructed offline using a representative validation dataset. Let $\bK_{\text{val}} \in \mathbb{R}^{n_{\text{val}}\times d}$ denote the set of key vectors extracted from the dataset, where $n_{\text{val}}$ is the number of tokens and $d$ is the embedding dimension.

The $d$-dimensional key embedding space is partitioned into $g$ disjoint subspaces (i.e., groups), each with dimensionality $\bar{d}=d/g$. Conceptually, this transforms a key vector $\bk \in \mathbb{R}^d$ into $g$ sub-vectors $\bar{\bk}_i \in \mathbb{R}^{\bar{d}}$. Within each of these $g$ groups, we apply a clustering method (e.g., K-means) to the corresponding sub-vectors across the validation dataset $\bK_{\text{val}}$ to derive a codebook of $c$ cluster centroids for each group. The complete codebook is denoted as $\bC \in \mathbb{R}^{g \times c \times \bar{d}}$. The generation process of $\bC$ is conceptually illustrated in the diagram below:

\begin{center}
\begin{tikzpicture}[node distance=2.4cm, every node/.style={font=\large}]
  \node (K)   {$\bK_{\text{val}} \in \mathbb{R}^{n_{\text{val}} \times d}$};
  \node (Kbar) [right=of K] {$\bar{\bK}_{\text{val}} \in \mathbb{R}^{g \times n_{\text{val}} \times \bar{d}}$};
  \node (C)   [right=of Kbar] {$\bC \in \mathbb{R}^{g \times c \times \bar{d}}$,};

  \draw[->, thick] (K) -- node[above, font=\small, yshift=2pt] {\textit{split into groups}} (Kbar);
  \draw[->, thick] (Kbar) -- node[above, font=\small, yshift=2pt] {\textit{quantization}} (C);
\end{tikzpicture}
\end{center}
where $n$ is the sequence length of LLMs.

\subsection{Heterogeneous Attention Computation}
\label{sec:hac}
Building upon the key cache quantization described in the previous section, we propose a heterogeneous attention computation framework to enable efficient inference with memory constraints while maintaining model accuracy.

Given key and value caches $\bK \in \mathbb{R}^{n \times d}$ and $\bV \in \mathbb{R}^{n \times d}$, the standard attention mechanism computes the output $\by$ for a query $\bq \in\mathbb{R}^d$ as
\begin{equation}
    \by =\operatorname{softmax}(\frac{\boldsymbol{q} \cdot \bK^\top}{\sqrt{d}})\bV = \operatorname{softmax}(\frac{\bz}{\sqrt{d}})\bV,
    % ,~~ where~\bz =\boldsymbol{q} \bK^T
\end{equation}
where $\bz = \bq \cdot \bK^\top$ represents the attention scores. 
% This computation exhibits $O(n^2d)$ complexity for processing a sequence of length $n$ and requires storing $\bK$ and $\bV$, posing significant memory and computational burdens, especially for long sequences during decoding.

% To mitigate these challenges, \ours~combines three core techniques: (1) approximate attention score computation using the quantized key cache representation, (2) dynamic token eviction based on approximate attention scores, and (3) offloading of the value cache ($\mathbf{V}$) and final computation to the CPU. The overall approximate attention computation pipeline involves steps executed on both the GPU and CPU.

\ours~combines two core approximate computation techniques, which reduces both memory requirements and computational complexity. The first is built upon the key quantization, the attention scores $\bz$ are estimated with quantized centroids of key matrix, denoted as $\widetilde{\bz} = \bq \cdot \mathcal{Q}(\bK)^\top$. The second approximation comes from the KV eviction, dynamically pruning less impactful tokens from the cache. Thus it enables adaptive memory usage across layers, denoted as $\mathcal{F}_{evict}(\widetilde{\ba})$. Combined these two approximate strategies with value offloading, the whole computation can be expressed as
\begin{equation}
\widetilde{\by}
\approx
\colorbox{myGreen}{%
  $\displaystyle
  \mathrm{softmax}\!\bigl(\tfrac{\bq\cdot \mathcal{Q}(\bK)^\top}{\sqrt{d}}\bigr)
  $%
}%
\;\,
\colorbox{myCyan}{%
  $\displaystyle \bV$%
}%
\;=\;
\colorbox{myGreen}{%
  $\displaystyle
  \mathrm{softmax}\!\bigl(\tfrac{\tilde \bz}{\sqrt{d}}\bigr)
  $%
}%
\;\,
\colorbox{myCyan}{%
  $\displaystyle \bV$%
}
\;\approx\;
\colorbox{myGreen}{%
  $\displaystyle
  \mathcal{F}_{evict}(\tilde \ba)
  $%
}%
\;\,
\colorbox{myCyan}{%
  $\displaystyle \mathcal{F}_{evict}(\bV)$%
}.
\label{eq:hcattn}
\end{equation}

\textbf{1) GPU part in Eq.~\ref{eq:hcattn}}

In the approximate attention score computation, the computation is performed group-wise. $\bK$ is divided into predefined groups. Each sub-group is represented as nearest neighbor of the centroids in the codebook, serves as a compressed representation of  $\bK$,  stored in GPU memory. The compressed representation is an index matrix  $\bP \in \mathbb{Z}^{n \times g}$, where $\bP_{i,j} \in \{1,\dots,c\}$. 

Similarly, query vector is partitioned into same groups.  Instead of computing similarity between query and key value, the approximation computing is performed between the query $\bar{\bq}$ and the codebook $\bC$, yields an intermediate hash table, denoted as $\bT=\bar{\bq} \cdot \bC$ where $\bT \in \mathbb{R}^{g \times c}$. The computational cost is $O(dc)$, constant with respect to sequence length $n$.

The approximation computation $\bq\cdot\mathcal{Q}(\bK)^\top$ is then obtained by aggregating the indexing result from each group, which is indexing $\bT$ using index saved in $\bP$:
\begin{equation}
   {\tilde{\bz}_{j}} =\sum_{i=1}^{g}{\bT}_{i,{\bP_{j,i}}},
\end{equation}
where $\widetilde{\bz}_j$ is the approximate attention score at the $j$-th of the sequence.
Approximate attention scores  $\widetilde{\ba}$ are then derived as  $\widetilde{\ba} = \operatorname{softmax}(\tilde{\bz}/\sqrt{d})$.

Different from existing methods, we argue that the  
token importance are context-aware and exhibits different distributions across layers. Following this observation, we propose an eviction criterion based on cumulative magnitude of attention scores, denoted as $\mathcal{F}_{evict}(\widetilde{\ba})$.

We first sort the sequence $\widetilde{\ba}$ in descending order. 
$\Pi_{n} = \{\pi_{1},\pi_{2},\cdots, \pi_{n}\}$ is ordered index . 
$\widetilde{\ba}  = (a_{\pi_{1}}, a_{\pi_{2}}, \dots, a_{\pi_{n}})$, where
$a_{\pi_{1}} \ge a_{\pi_{2}} \ge \cdots \ge a_{\pi_{n}}$.
Given a threshold $\tau \in (0,1]$, the selected index $k^*$ is defined by
\begin{equation}
k^* \;=\; \min\Bigl\{\,k \in \{1,2,\dots,n\} \;\Bigm|\; \sum_{i=1}^{k} a_{\pi_i} \;\ge\; \tau \Bigr\}.
\label{eq:eviction}
\end{equation}
The attention computation can be approximated by a weighted sum of a selected subset of values, as the cumulative contribution of a few top-scoring entries dominates the output $\by$. We denote the approximate attention scores as $\widetilde{\ba}^* = \{a_{(1)}, a_{(2)}, \dots, a_{(k^)}\}$, and the corresponding selected indices as $\Pi_{k^*} = \{\pi_1, \pi_2, \dots, \pi_{k^*}\}$. Note that both sequences are transferred to the CPU to participate in the final output computation with value offloading.

% In words, $k^*$ is the smallest number of top probabilities whose cumulative sum meets or exceeds the threshold $\tau$.

% \begin{equation}
% \mathcal{S}(\widetilde{a}, \tau) = \min \left\{ k \in \{1, \dots, n\} \;\middle|\; \sum_{i=1}^{k} \widetilde{a}_{\pi(i)} \geq \tau \right\}
% \end{equation}

% \begin{equation}
% I_{select} = \left\{\pi_j \middle| j\in \{ 1,2,\cdots,\mathcal{S}(\widetilde{a}, \tau)\} \right \}
% \end{equation}

\textbf{2) CPU part in Eq.~\ref{eq:hcattn}}

Based on approximate key cache eviction, we propose fully offloading the value matrix $\bV\in \mathbb{R}^{n\times d}$ to CPU memory to optimize the computation process. The weighted sum of values using selected keys can be represented as:
\begin{equation}
    \widetilde{\by} = \ba\bV \approx \mathcal{F}_{evict}(\widetilde{\ba})\mathcal{F}_{evict}(\bV) = {\widetilde{\ba}^*}\bV^* = \sum_{i \in \Pi_{k^*}} \widetilde{\ba}^*_{i}\times \bV_i.
\end{equation}

\subsection{Overhead and Efficiency Analysis}
\textbf{Compression ratio.} The memory savings stem from a combination of value offloading and key quantization. Intuitively, offloading the value matrix to CPU memory releases half of the memory budget, since the key and value matrices have equivalent dimensions. The other strategy of  memory reduction is achieved by caching only the indices of quantized codebooks. As described in Sec.~\ref{sec:hac}, given a group number $g=64$ (or $g=32$) and embedding dimension $d=128$, the key index matrix $\bP\in \mathbb{Z}^{n\times g}$ occupies only a fraction of the original memory,  with an effective footprint of $g/d=1/2$ (or $1/4$), respectively. The overall memory budget breakdown is illustrated in Table ~\ref{tab:mem_budget}.
% \begin{center}
% \begin{tikzpicture}[node distance=2.4cm, every node/.style={font=\small}]
%   \node (input)   {$\text{softmax}(\frac{q\cdot K^\top}{\sqrt{d}})V$};
%   \node (voffload) [right=of input] {50\% memory budget};
%   \node (g64)   [above right=0.1cm and 2.5cm of voffload] {25\% memory budget};
%   \node (g32)   [below right=0.1cm and 2.5cm of voffload] {12.5\% memory budget};

%   \draw[->, thick] (input) -- node[above, font=\small, yshift=1.5pt] {\textit{value offloading}} (voffload);
%   \draw[->, thick] (voffload) -- node[above, font=\small, yshift=2pt] {$g=64$} (g64);
%   \draw[->, thick] (voffload) -- node[below, font=\small, yshift=2pt] {$g=32$} (g32);
% \end{tikzpicture}
% \end{center}

\textbf{Computational cost.}
% \begin{wrapfigure}{r}{0.4\textwidth}
%     \centering
%     \includegraphics[width=0.38\textwidth]{float_ops_comparison.png}
%     \caption{A comparison of floating-point operation (FLOP) counts. 
%     \label{fig:ops}
% \end{wrapfigure}
The reduction in computational cost arises from the approximate computation of attention scores, denoted as $ \bq \cdot \mathcal{Q}(\bK)^\top$.  Rather than evaluating attention scores against all keys, the computation is restricted to the cluster centers in the codebook, significantly lowering the number of operations, as shown in Table ~\ref{tab:comp_complexity}.
% The computational cost under this approximation can be summarized as follows:

% \setlist{leftmargin=*}
% \begin{itemize}
% \item \textbf{Multiplications}: Reduced from $\mathcal{O}(n^2 d)$ to $\mathcal{O}(n d c)$
% \item \textbf{Additions}: Reduced from $\mathcal{O}(n^2 d)$ to $\mathcal{O}(n^2 g)$
% \end{itemize}

\begin{table}[h]
\centering
\scriptsize  % 使用更小的字体
\setlength{\tabcolsep}{4.5pt} 
\caption{Comparison of memory compression ratio and computational cost }
\begin{subtable}[t]{0.48\linewidth}
\centering
\caption{Compression ratio of GPU}
\label{tab:mem_budget}
\begin{tabular}{llll}
\toprule
\textbf{Strategy}               & $\bK$~\textbf{Budget} &  $\bV$~\textbf{Budget} &  \textbf{Total} \\
\midrule             
Full-attention                      & 100\%             & 100\%              & 100\%                   \\
Value Offloading (VO)          & 100\%             & 0\%                & 50\%                    \\
VO + quantization ($g=64$)         & 50\%              & 0\%                & 25\%                    \\
VO + quantization ($g=32$)         & 25\%             & 0\%                 & 12.5\%                  \\
\bottomrule
\end{tabular}
\hfill % 水平间距
\end{subtable}
\begin{subtable}[t]{0.48\linewidth}
\centering
\hfill  % 水平间距
\caption{Computational cost}
\label{tab:comp_complexity}
\begin{tabular}{lcc}
\toprule
\textbf{Operation} & \textbf{Original} & \textbf{\ours} \\
\midrule
Multiplications & $O(n^{2}d)$ & $O(ndc)$ \\
Additions       & $O(n^{2}d)$ & $O(n^{2}g)$ \\
\bottomrule
\end{tabular}
\end{subtable}

\end{table}

% Here, $n$ denotes the sequence length, $d$ the embedding dimension, $c$ the number of cluster centers, and $g$ the number of codebook groups. 
The cost reduction is particularly significant in practice since $c \ll n$ in the long context applications.
% Fig.~\ref{fig:combined_plots}(a) compares the operation counts between standard attention and two typical configurations of Fast $\mathbf{q}\mathbf{K}^\top$.
% The results demonstrate that for sequences up to 1M tokens, Fast $\mathbf{q}\mathbf{K}^\top$ requires less than 1/5th the operations of the original approach.

\textbf{Communication overhead.} Thanks to KV cache eviction, only a small subset of attention scores needs to be transferred to CPU memory (see Sec.~\ref{sec:hac}). Given a threshold $\tau$, if 20\% of important keys are retained across all layers of the LLM, the size of the resulting tensor $\widetilde{\ba}^*$ can be estimated as $\frac{1}{5} \times n \times L \times H \times 2~\text{Bytes}$, where  $L$ is the number of attention layers and $H$ denotes the number of attention heads. For example, using Llama3-8B, which has $L = 32$ layers and $H = 8$ attention heads per layer, and assuming a sequence length of $n = 10^6$, the total communication overhead amounts to approximately 102.4 MB. This is negligible compared to the bandwidth capacity of a standard PCIe connection. This demonstrates that HCAttention's sparsified offloading mechanism can scale efficiently.

% \begin{figure}
%     \centering
%     \includegraphics[width=1 \linewidth]{combined_plots.png}
%     \caption{(a)A comparison of floating-point operation (FLOP) counts. 
% Here, $g$ denotes the number of groups, and $b$ represents $2^b$ vectors per group.(b)Our method outperforms existing approaches in both performance and memory usage, achieving high accuracy even with ultra-high compression rates.}
%     \label{fig:combined_plots}
% \end{figure}

\section{Experiment Results}
In this section, we evaluate \ours~on long-context benchmarks using Llama models. We further conduct ablation studies to analyze the individual contribution of each component.
% The reported results demonstrate that \ours~significantly reduces GPU memory budget while maintaining performance on various tasks.
% We first introduce the experimental setup, including hardware configuration, software environment, evaluation metrics, datasets, and baseline methods. Subsequently, we present detailed empirical results and discussions. Finally, we verify the effectiveness of each component in our solution through ablation experiments.

\subsection{Setup}
\textbf{Models and datasets.} We employ two state-of-the-art open-source models for our experiments: Llama-2-7B-32K-Instruct~\cite{touvron2023llama} and Llama-3-8B-Instruct-Gradient-1048k~\cite{grattafiori2024llama}.
For long-context evaluation, we use two widely adopted benchmarks: LongBench~\cite{bai2024longbench} and the Needle-in-a-Haystack (NIAH) benchmark~\cite{needleinahaystack}.
%for short-context evaluation, we compare the performance on MMLU~\cite{hendryckstest2021}, MBPP~\cite{austin2021program}, and MT-Bench~\cite{bai2024mt}.

\textbf{Baseline methods.} We compare \ours~against the original model inference without compression (referred as Full) and several state-of-the-art KV cache compression algorithms: H2O~\cite{zhang2023h2o}, TOVA~\cite{oren2024transformers}, DuoAttention (referred as Duo)~\cite{xiao2024duoattention},  and StreamingLLM (referred as SLLM )~\cite{xiao2023efficient}.
% StarAttention~\cite{acharya2024starattention}
We follow the baseline configurations as described in their respective papers. Specifically, SLLM employs 64 sink tokens and retains 50\% of the sequence as recent tokens. DuoAttention adopts the same configuration for its streaming heads, while the retrieval heads are selected proportionally.
 
\textbf{Implementation details.} 
In the experiments, we set the attention sparsification threshold to $\tau = 0.9$. Different memory budgets were explored: 50\% memory used only value offloading (without key quantization); 25\% memory combined value offloading with key quantization at $g=64$; and the 12.5\% memory budget used key quantization at $g=32$.
The codebook is generated using the MiniBatchKMeans algorithm~\cite{sculley2010web} implemented in Scikit-learn~\cite{pedregosa2011scikit}. For the clustering process, we used 3 randomly selected samples with a maximum of 200 iterations and a batch size of 10,000.

\subsection{Long Context Benchmarks}
\begin{table}[htbp]
\centering
\caption{LongBench performance of different methods under varying memory budgets. Results show that the \ours~generally maintains strong performance, achieving competitive or superior results, particularly under constrained memory budgets (25\% and 12.5\% KV cache).}
\label{tab:longbenchlargetable}
\resizebox{\textwidth}{!}{
\begin{tabular}{c*{8}{c}@{\hspace{3em}}c*{8}{c}}
\toprule
\multirow{2}{*}{\textbf{Dataset}} & \multicolumn{6}{c}{\textbf{Llama-3-8B-Instruct-1048K(50\%)}} & 25\% &12.5\% & \multicolumn{6}{c}{\textbf{Llama-2-7B-Instruct-32K(25\%)}} & 50\% &12.5\%\\ \cmidrule(lr){2-17}
 & \textbf{Full} & \textbf{H2O} & \textbf{SLLM} & \textbf{TOVA} & \textbf{Duo} & \textbf{Ours} & \textbf{Ours} & \textbf{Ours} & \textbf{Full} & \textbf{H2O} & \textbf{SLLM} & \textbf{TOVA} & \textbf{Duo} & \textbf{Ours} & \textbf{Ours}& \textbf{Ours}\\
\midrule
\textbf{Average}     & 43.2 & 39.6 & 37.2 & 39.7 & \textbf{44.0} & 43.2  & 43.2 & 42.5   & 42.0 & 32.5 & 33.2 & 34.8 & 38.8 & \textbf{41.5} & 42.0 & 40.0 \\
\hline                                                                                  
\multicolumn{1}{c}{\textbf{Single-Document QA}} \\                                      
NarrativeQA          & 26.6 & 25.1 & 22.1 & 25.6 & 24.5 & \textbf{26.0}  & 26.1 & 24.3   & 24.1 & 19.1 & 21.1 & 23.1 & 20.5 & \textbf{24.5} & 24.8 & 25.0  \\
Qasper               & 29.2 & 20.7 & 21.8 & 23.1 & 26.9 & \textbf{29.8}  & 29.7 & 27.9   & 33.2 & 16.8 & 17.7 & 20.9 & 26.6 & \textbf{31.4} & 32.4 & 32.9  \\
MultiFieldQA-en      & 52.6 & 38.5 & 28.1 & 44.9 & 51.4 & \textbf{53.9}  & 52.7 & 50.5   & 34.0 & 21.0 & 16.7 & 18.2 & 25.5 & \textbf{34.3} & 35.0 & 33.8  \\
%MultiFieldQA-zh      & 50.6 & 38.3 & 31.1 & 40.8 & \textbf{52.4} & 51.3  & 51.1 & 36.1   & 45.8 & 19.8 & 22.5 & 25.0 & 39.2 & \textbf{46.7} & 46.7 & 34.3  \\
\hline                                                                                                                                                      
\multicolumn{1}{c}{\textbf{Multi-Document QA}} \\                                                                                                           
HotpotQA             & 40.4 & 36.8 & 39.3 & 38.5 & \textbf{41.6} & 39.7  & 40.3 & 41.6   & 48.0 & 39.6 & 40.4 & 47.5 & \textbf{50.4} & 46.8 & 48.0 & 46.0  \\
2WikiMQ              & 28.8 & 28.0 & 29.2 & 26.9 & \textbf{29.1} & 28.8  & 29.9 & 28.7   & 35.6 & 28.9 & 29.7 & 31.2 & 33.4 & \textbf{34.9} & 36.2 & 33.9  \\
Musique              & 24.2 & 19.2 & 20.5 & 23.1 & 24.7 & \textbf{25.1}  & 25.1 & 24.6   & 23.0 & 20.6 & 20.1 & 21.0 & 19.3 & \textbf{23.7} & 23.5 & 22.9  \\               
%DuReader (zh)        & 30.4 & 24.9 & 9.4  & 27.0 & 29.3 & \textbf{29.7}  & 30.3 & 28.5   & 25.1 & 15.6 & 14.0 & 15.5 & 24.0 & \textbf{26.2} & 25.4 & 20.3  \\
\hline                                                                                                                                                      
\multicolumn{1}{c}{\textbf{Summarization}} \\                                                                                                              
GovReport            & 34.2 & 29.4 & 29.1 & 30.1 & 32.7 & \textbf{34.5}  & 34.3 & 33.5   & 31.2 & 20.7 & 24.1 & 22.9 & 28.0 & \textbf{30.8} & 30.8 & 28.9  \\
QMSum                & 24.5 & 22.9 & 22.1 & 23.2 & \textbf{24.2} & 24.1  & 24.1 & 23.9   & 20.8 & 18.9 & 20.1 & 20.2 & \textbf{21.5} & 20.7 & 20.7 & 20.3  \\
MultiNews            & 27.7 & 25.5 & 24.9 & 26.3 & \textbf{27.7} & 27.5  & 28.2 & 27.5   & 27.1 & 19.2 & 20.5 & 21.4 & 25.0 & \textbf{26.8} & 27.1 & 26.0  \\
%VCSUM (zh)           & 11.4 & 13.5 & 12.1 & 11.6 & 10.5 & \textbf{12.5}  & 13.5 & 15.7   & 14.5 & 10.7 & 14.4 & 11.9 & 12.4 & \textbf{15.1} & 14.8 & 12.1  \\
\hline                                                                                                                                                     
\multicolumn{1}{c}{\textbf{Few-shot Learning}} \\                                                                                                          
TREC                 & 71.5 & 64.0 & 67.0 & 54.0 & 71.0 & \textbf{71.5}  & 71.5 & 72.0   & 71.5 & 48.5 & 56.5 & 47.0 & 68.5 & \textbf{72.0} & 72.0 & 70.0  \\
TriviaQA             & 87.7 & 86.0 & 86.1 & 85.0 & 87.1 & \textbf{87.2}  & 87.2 & 86.8   & 86.2 & 85.2 & 85.2 & 85.7 & 86.2 & \textbf{86.8} & 86.9 & 82.1  \\
SAMSum               & 42.5 & 40.8 & 40.3 & 40.5 & 41.8 & \textbf{42.4}  & 42.8 & 42.8   & 42.1 & 39.7 & 37.4 & 36.2 & 33.1 & \textbf{41.5} & 41.1 & 39.5  \\
%LSHT (zh)            & 38.0 & 25.0 & 25.5 & 24.5 & 30.0 & \textbf{37.5}  & 37.5 & 35.5   & 34.5 & 16.5 & 17.5 & 18.5 & 25.5 & \textbf{34.0} & 34.0  & 24.0  \\
\hline                                                                                                                                                      
\multicolumn{1}{c}{\textbf{Synthetic Task}} \\                                                                                                              
% Passage Count        & 1.0  & 2.1  & 1.6  & 1.0  & 0.0  & \textbf{1.0}   & 1.0  & 1.0    & 0.0  & 0.5  & 0.6  & 0.0  & \textbf{0.3}  & 0.0  & 0.0  & 0.7   \\  
Passage Retrieval-en & 81.0 & 74.8 & 49.0 & 72.0 & \textbf{87.0} & 80.1  & 80.5 & 80.0   & 50.9 & 19.5 & 19.1 & 30.2 & \textbf{47.3} & 46.5 & 47.9 & 41.9  \\
%Passage Retrieval-zh & 62.2 & 52.6 & 38.9 & 46.1 & \textbf{62.2} & 57.0  & 62.2 & 38.0   & 37.7 & 11.8 & 16.8 & 32.4 & \textbf{40.9} & 38.7 & 37.1 & 23.6  \\
\hline                                                                                                                                                      
\multicolumn{1}{c}{\textbf{Code Completion}} \\                                                                                                             
LCC                  & 38.2 & 43.1 & 41.9 & 42.3 & \textbf{44.2} & 39.2  & 37.8 & 36.1   & 51.2 & 45.8 & 44.3 & 47.9 & 48.3 & \textbf{51.7} & 52.0 & 50.5  \\                 
RepoBench-P          & 38.9 & 40.0 & 37.6 & 40.1 & \textbf{46.1} & 39.8  & 38.8 & 36.8   & 51.6 & 45.2 & 45.3 & 49.0 & 48.6 & \textbf{50.9} & 51.6 & 46.9  \\
\bottomrule
\end{tabular}
}
\end{table}

\textbf{LongBench} is a specialized benchmark for evaluating long-context comprehension capabilities of LLMs, consisting of various tasks including single-document QA, multi-document QA, summarization, code understanding, and logical reasoning. We presents a comprehensive evaluation of different methods on the LongBench across various memory constraints, utilizing two different Llama models. The memory budgets are represented as percentages of the full KV cache size (Full, 50\%, 25\%, 12.5\%).

As shown in Table~\ref{tab:longbenchlargetable}, \ours~consistently demonstrates robust performance. In many tasks and across both model sizes, \ours~is competitive with or surpasses other methods under these memory constraints. It achieves performance nearly indistinguishable from the full attention baseline under a 25\% GPU memory budget for both models. Remarkably, it stays competitive with only 12.5\% of the cache with Lllma-3-8B-Instruct-1048, setting a new state-of-the-art in LLM KV cache compression.  This highlights the robustness of our heterogeneous attention design, especially in balancing GPU memory efficiency with attention quality.

These results collectively demonstrate that \ours~enables practical long-context inference with minimal trade-offs in accuracy, offering a scalable and generalizable solution for memory-efficient deployment of LLMs.

\textbf{NIAH} primarily measures a model's ability to precisely locate and extract key information from extremely long contexts. We maintains the same experimental settings to those used in LongBench. 
% The StreamingLLM scheme employs 64 sink tokens, with 50\% of the sequence length allocated to recent tokens. The Streaming Heads in the DuoAttention adopt the same approach as the SLLM scheme, while the Retrieval Heads are selected proportionally. 
We investigate the impact of different "needles" on the task performance. We take ``\textit{The best thing to do in Paris is buy a fresh croissant and lounge by the Seine at twilight.}'' as an example ``needle''. 
% \textcolor{red}{TODO: add other Figure 2.}\\
% \textcolor{red}{TODO:What is config of DuoAttention under 12.5\%}

\begin{figure}[htbp]
    \centering
    \includegraphics[width=1\linewidth]{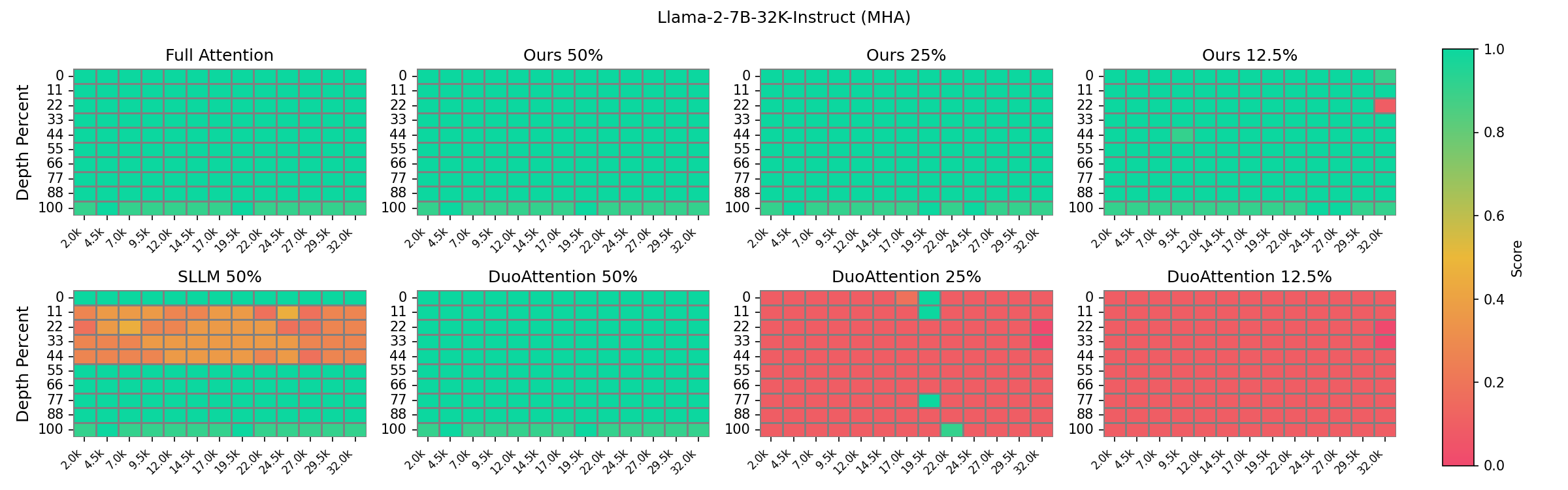}
    \caption{DuoAttention can maintain model performance at 50\% memory budget, but lower budgets cause model collapse. Our solution achieves performance close to the full attention model at 50\% and  25\% memory budget, and crucially maintains model effectiveness even at lower budgets of 12.5\%, significantly outperforming DuoAttention in these constrained scenarios.}
    \label{fig:NIAH32K}
\end{figure}

We conduct NIAH testing on Llama-2-7B-32K-Instruct. As shown in Fig.~\ref{fig:NIAH32K}, DuoAttention maintains model performance at a 50\% memory budget but fails under lower budgets, leading to performance collapse. In contrast, our method achieves performance comparable to the full attention model at both 50\% and 25\% memory budgets, and crucially, it remains effective even at the extreme 12.5\% budget significantly outperforming DuoAttention under such constrained settings. Additional results with alternative ``needle'' queries are provided in the Appendix.

\textbf{Efficiency result}.
By combining the Llama-3-8B model (using FlashAttention2~\cite{dao2023flashattention2} for 1024k window prefilling) with our approach, which integrates $g=32$ key quantization and value offloading, we successfully processed sequences up to 4 million tokens on a single A100 GPU (80GB memory).

\subsection{Ablation Studies}
\subsubsection{Ablation on KV eviction}
\begin{figure}[htbp]
    \centering % 整体居中
    \begin{minipage}[c]{0.48\textwidth}
     \centering
     \tiny 
     \captionof{table}{Performance under varying KV eviction thresholds.} 
     \label{tab:threshold}
     \begin{tabular}{lrrrrrr}
     \toprule
     \textbf{Threshold $\tau$}        & 0.3  & 0.5  & 0.7  & 0.9   & 1    \\
     \textbf{Selection Ratio \%}        & 1.9 & 3.2 & 6.0 & 15.6 & 100  \\
     \midrule
     Average                 & 40.2  & 41.2  & 43.9  & 47.2  & 46.9   \\
     2WikiQA                 & 27.0  & 29.6  & 33.6  & 36.0  & 35.6    \\
     GovReport               & 29.2  & 30.3  & 31.0  & 30.2  & 31.2    \\
     HotpotQA                & 35.3  & 38.6  & 44.2  & 48.7  & 48.0    \\
     LCC                     & 45.4  & 48.6  & 51.4  & 52.0  & 51.5    \\
     MultiNews               & 25.9  & 25.2  & 26.4  & 26.6  & 27.1    \\
     MultiFieldQA-en         & 28.8  & 25.3  & 29.8  & 36.8  & 34.0    \\
     PassageRetrieval-en     & 46.8  & 41.4  & 40.9  & 51.2  & 50.9    \\
     Qasper                  & 23.7  & 24.4  & 28.5  & 32.7  & 33.2    \\
     RepoBench-P             & 45.1  & 48.3  & 49.8  & 51.2  & 51.6    \\
     SAMSum                  & 38.9  & 41.2  & 41.1  & 41.8  & 42.1     \\
     TREC                    & 68.0  & 68.0  & 70.1  & 72.0  & 71.5     \\
     TriviaQA                & 68.8  & 73.1  & 80.1  & 86.9  & 86.2    \\
     \bottomrule
     \end{tabular}
    \end{minipage}
    \hfill % 在两个 minipage 之间填充水平空间
    \begin{minipage}[c]{0.48\textwidth}
        \centering
        \includegraphics[width=0.95\linewidth]{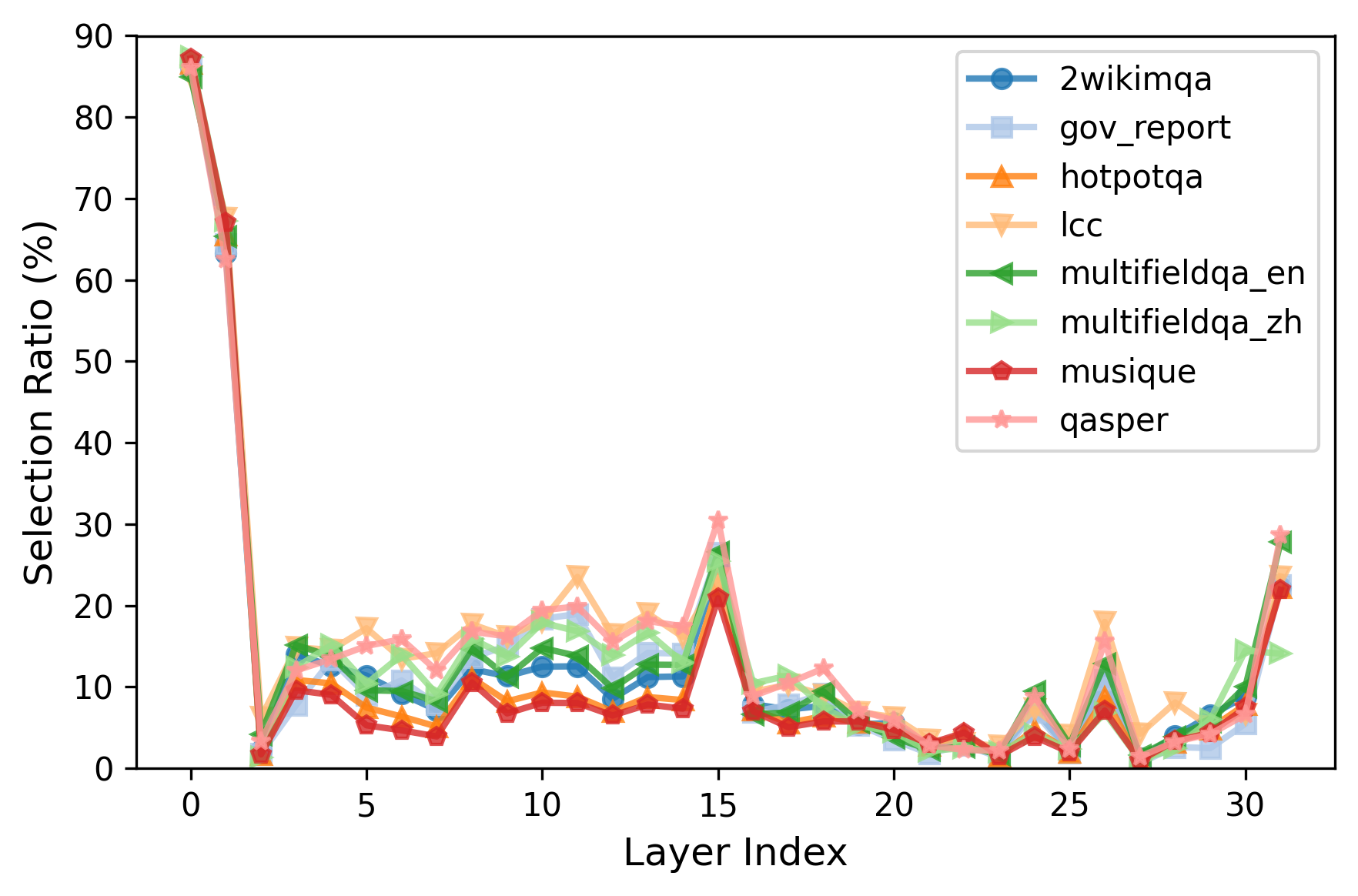}
        \captionof{figure}{Comparison of dynamic KV eviction selection ratio across transformer layers.}
        \label{fig:Layer_wise_selection_ratio}
    \end{minipage}
    % 注意：如果两个 minipage 宽度之和接近或等于 \textwidth，
    % 并且它们之间没有 \hfill 或其他水平间距，可能会导致换行。
    % 确保 minipage 的总宽度加上间距小于或等于 \textwidth。
\end{figure}

% Based on experiments with the Llama2-7B-32K architecture on the LongBench benchmark, we systematically evaluated the impact of different attention accumulation thresholds on KV cache utilization efficiency. The experiment employed a grouping strategy of 64 clusters, using an 8196$\times$2 dimensional quantization center table to compress Key vectors from 128$\times$16 (BF16 format) to 64$\times$16 (int16) format, equivalent to 8-bit quantization.
We evaluate the effectiveness of applying layer-wise dynamic KV eviction (i.e., attention score thresholding) without key quantization or value offloading. This ablation study is conducted using the Llama-2-7B-32K model on the LongBench benchmark. 

% As shown in Table~\ref{tab:threshold}, we vary the threshold $\tau$ from 0.3 to 1.0. Results indicate that setting $\tau=0.9$ allows the system to prune over 80\% of tokens while maintaining the accuracy on  average. Lower thresholds include less tokens but introduce noise and diminish efficiency, leading to performance degradation. Therefore, $\tau = 0.9$ strikes a favorable balance between information retention and computational efficiency.

As shown in Table~\ref{tab:threshold}, we varied the threshold $\tau$  from 0.3 to 1.0. Results indicate that setting $\tau=0.9$ allows the system to prune over 80\% of tokens while maintaining accuracy on average. Lower thresholds, while pruning more tokens (i.e., retaining fewer), introduce noise (or discard essential information) and lead to performance degradation. Therefore, $\tau=0.9$~strikes a favorable balance between information retention and computational efficiency.

Layer-wise token selection ratios (Figure~\ref{fig:Layer_wise_selection_ratio}) demonstrate that dynamic KV eviction benefits from adapting to each layer's redundancy. Early layers permit aggressive eviction, middle layers require caution, and the final layer varies by task, highlighting the efficiency of a layer-aware approach for memory management.

\subsubsection{Key quantization quality}
\begin{figure}[htbp]
    \centering
    \includegraphics[width=1\linewidth]{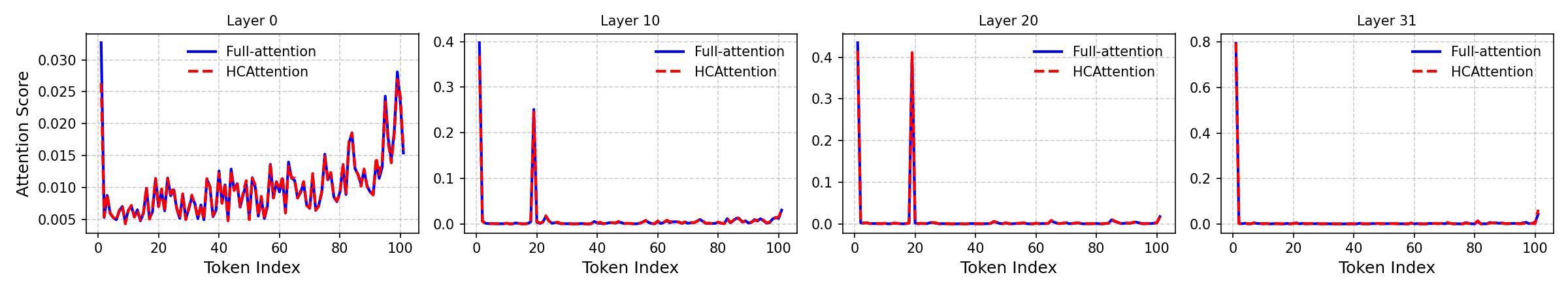}
    \caption{Attention scores in different layers between full-attention and group quantization ($g=32$, $c=8192$).}
    \label{fig:attention-diff}
\end{figure}
We analyze the quantization quality in \ours~by comparing the output attention scores between the full-attention and quantized attention. As shown in Figure~\ref{fig:attention-diff}, we plot the attention scores for the first few hundred tokens, with full attention in blue and quantized attention in red. The two curves largely overlap, indicating that the quantization introduces negligible loss in attention scores. Furthermore, we visualize four different layers and observe that even in the final layer, there is no noticeable accumulation of error, demonstrating the effectiveness and stability of our key quantization.

\subsubsection{Key quantization strategy comparison}
To assess the impact of quantization settings, we conduct ablation studies on (a) the number of codebook centroids and (b) the group size in channel-wise partitioning, with KV eviction threshold $\tau=1$ to isolate the quantization effect. 

The results are summarized in Table~\ref{tab:ablstrategy}.Varying the number of centroids (256–8192) with a fixed group size ($g=32$), we observe a clear trade-off between quantization granularity and attention accuracy. Performance improves significantly from 256 to 2048 centroids and stabilizes at 4096+, highlighting the need for expressive codebooks to preserve semantic similarity.

For group size ($g$ = 16, 32, 64), larger sizes yield better performance by reducing quantization noise. Additionally, using separate codebooks for each group (Table~\ref{tab:ablcq}, last column) slightly increases memory but consistently improves accuracy, suggesting that decoupled clustering better captures diverse semantics in long contexts.

% We analyze different group sizes ($g$ = 16, 32, 64). Overly aggressive compression introduces excessive quantization noise, resulting the lowest performance. As we increase the group size from 16 to 32 and 64, performance gradually recovers, indicating that finer subspace partitioning improves representational fidelity.
% Moreover, we test the performance without sharing codebook between the groups (the last coloumn in Table~\ref{tab:ablcq}).  While this slightly increases codebook memory, it consistently yields marginal accuracy gains over shared strategy. This suggests that decoupled clustering across groups better captures diverse semantic structures in long-context sequences.

\begin{table}
    \centering
    \scriptsize  % 使用更小的字体
    \setlength{\tabcolsep}{4.5pt}  
    \caption{Ablation on key quantization settings (``$*$'' indicates that the groups do not share the codebook, and ``-'' denotes the full-attention baseline.)}
    \label{tab:ablstrategy}
    \begin{subtable}[t]{0.48\linewidth}
    \caption{Effect of codebook granularity}
    \label{tab:ablthresh}
        \begin{tabular}{lrrrrrr}
             \toprule
             \textbf{number of centers}        & 256     & 1024  & 2048   & 4096   & 8192    \\
             % \textbf{EqualBits}      & 2       & 4     & 4      & 4      & 4  \\
             \midrule
             Average                 & 23.8  & 36.4  & 41.4  & 42.2  & 43.5  \\
             2WikiQA                 & 23.8  & 30.9  & 33.6  & 33.7  & 32.8  \\
             GovReport               & 9.5   & 22.8  & 25.6  & 26.1  & 28.7  \\
             HotpotQA                & 29.4  & 37.7  & 45.4  & 45.0  & 45.1  \\
             LCC                     & 23.1  & 37.6  & 40.9  & 48.4  & 48.9  \\
             MultiNews               & 12.9  & 16.8  & 21.1  & 25.5  & 25.4  \\
             MultiFieldQA-en         & 19.3  & 31.2  & 34.0  & 31.6  & 34.3  \\
             PassageRetrieval-en     & 13.3  & 24.0  & 35.9  & 39.3  & 41.0  \\
             Qasper                  & 19.9  & 26.8  & 31.4  & 29.5  & 32.4  \\
             RepoBench-P             & 18.4  & 38.4  & 44.0  & 44.1  & 45.9  \\
             SAMSum                  & 18.1  & 34.0  & 36.6  & 38.2  & 37.5  \\
             TREC                    & 48.0  & 64.5  & 66.0  & 66.0  & 68.0  \\
             TriviaQA                & 50.3  & 72.6  & 81.9  & 78.6  & 81.9  \\
            \bottomrule
            \end{tabular}
    \end{subtable}
    \begin{subtable}[t]{0.48\linewidth}
    \caption{Effect of group dimension}
    \label{tab:ablcq}
        \begin{tabular}{lccccccc}
            \toprule                           
            \textbf{group size}      & -     & 16   & 32   & 64  & $64^*$\\
            % \textbf{EqualBits}  & 16    & 2    & 4    & 8    & 4    \\
            \midrule                           
            Average             & 46.9 & 12.8 & 43.5  & 46.5 & 45.6 \\ %43.0
            2WikiQA             & 35.6 & 18.0 & 32.8  & 35.2 & 35.5 \\ %32.8
            GovReport           & 31.2 & 2.1  & 28.7  & 31.1 & 31.0 \\ %27.9
            HotpotQA            & 48.0 & 27.0 & 45.1  & 46.9 & 46.4 \\ %47.5
            LCC                 & 51.5 & 4.8  & 48.9  & 51.3 & 50.3 \\ %49.3
            MultiNews           & 27.1 & 0.1  & 25.4  & 26.7 & 26.2 \\ %24.7
            MultiFieldQA-en              & 34.0 & 10.8 & 34.3  & 34.9 & 33.7 \\ %34.7
           PassageRetrieval-en         & 50.9 & 7.0  & 41.0  & 48.5 & 47.8 \\ %36.3
            Qasper              & 33.2 & 14.3 & 32.4  & 32.1 & 32.3 \\ %34.1
            RepoBench-P            & 51.6 & 7.3  & 45.9  & 51.5 & 48.4 \\ %46.4
            SAMSum              & 42.1 & 6.4  & 37.5  & 41.4 & 39.8 \\ %37.5
            TREC                & 71.5 & 28.5 & 68.0  & 72.0 & 69.5 \\ %65.0
            TriviaQA            & 86.2 & 27.7 & 81.9  & 86.7 & 86.0 \\ %80.1
            \bottomrule                        
        \end{tabular}
    \end{subtable}
\end{table}

\section{Conclusion}
In this work, we present \ours, a heterogeneous approximate attention framework that effectively addresses the KV cache memory bottleneck in long-context LLM inference. By combining key quantization, value offloading to CPU memory, and layer-wise KV eviction, \ours~significantly reduces GPU memory consumption while maintaining high-quality outputs and low inference latency. The framework is fine-tuning-free and can be readily integrated into standard Transformer inference pipelines. Empirical results on LongBench and NIAH benchmarks demonstrate that \ours~achieves state-of-the-art compression efficiency, and remains robust even under extreme memory constraints (as low as 12.5\% of the original KV cache size). Ablation studies further validate the effectiveness of our quantization strategy and attention sparsification design.  To the best of our knowledge, \ours~is the first to extend the Llama-3-8B model to process 4 million tokens on a single A100 GPU with 80GB memory.

For future work, two main directions appear particularly promising. Firstly, exploring the integration of \ours~with the DeepSeek MLA (Multi-head Latent Attention) presents a promising direction. Since MLA is inherent memory efficiency compared to MHA and GQA, combining it with \ours~holds the potential for even greater memory savings. Validating the effectiveness of this synergy will be a key focus of future research. Secondly, we aim to investigate the application of \ours~in broader scenarios, including multimodal models, AI Agent and Retrieval-Augmented Generation system. The efficient handling of long contexts offered by \ours~could unlock significant improvements in the performance and capabilities of these systems.

\printbibliography[heading=bibnumbered]  % 生成参考文献列表

\cleardoublepage % Ensures the appendix starts on an odd-numbered page (for two-sided documents)
\appendix % This command changes the numbering style for chapters/sections and resets counters for figures and tables.
\addcontentsline{toc}{chapter}{Appendix} % Adds "Appendix" to the Table of Contents if you're not using \chapter*
\setcounter{figure}{0} % Resets the figure counter to 0
\setcounter{table}{0}
\section{Alternative experiment results on NIAH}  
\begin{figure}[h]
    \centering
    \includegraphics[width=1\linewidth]{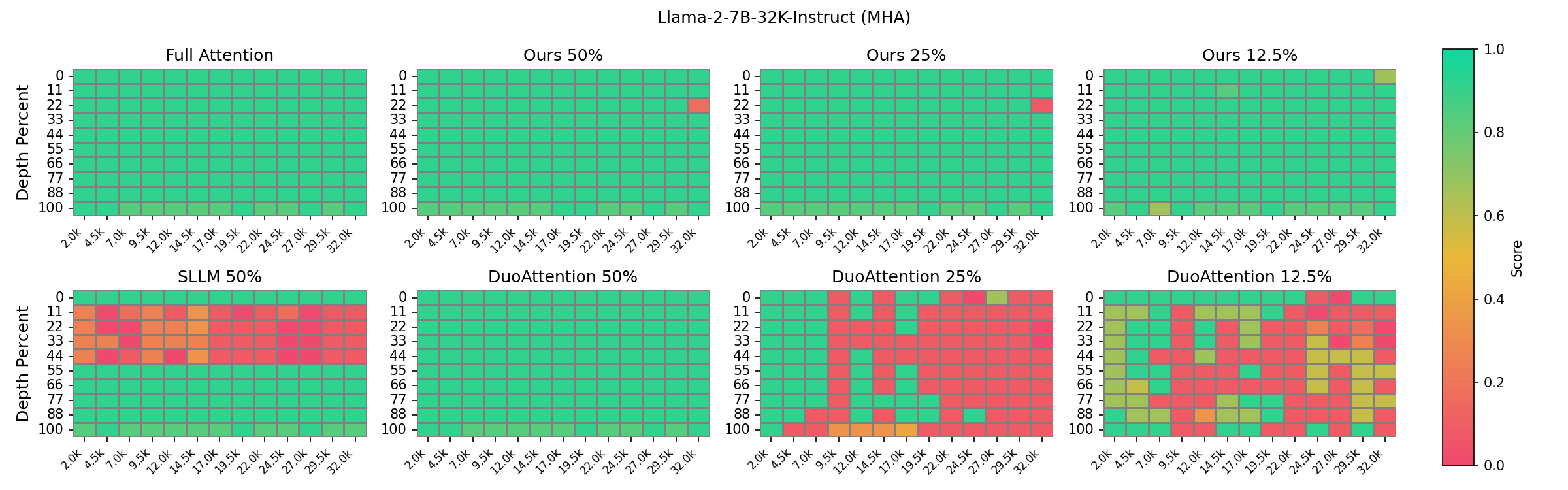}
    \caption{Visualization of NIAH test. The "needle" text is "Remember, the best thing to do in San Francisco is eat a sandwich and sit in Dolores Park on a sunny day.".}
    \label{fig:Needle-San-Francisco}
\end{figure}

\begin{figure}[h]
    \centering
    \includegraphics[width=1\linewidth]{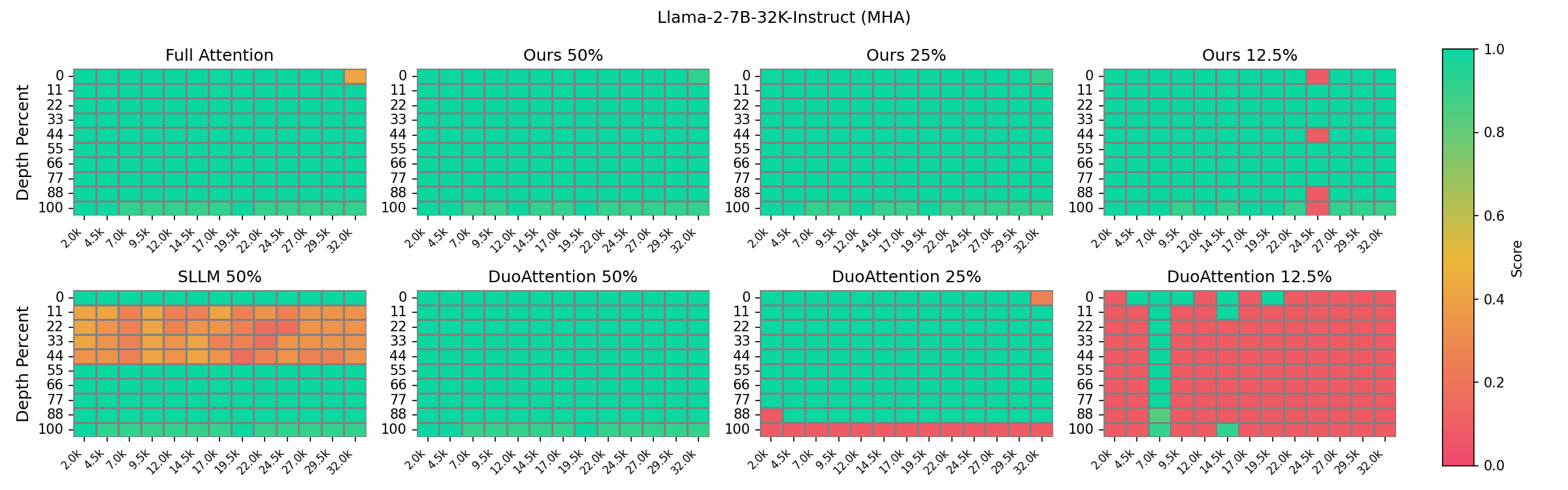}
    \caption{Visualization of NIAH test. The "needle" text is  "Remember, the best thing to do in New York is grab a bagel and people-watch in Washington Square Park at golden hour.".}
    \label{fig:Needle-New-York}
\end{figure}

This section presents the visualization results of NIAH test with alternative "needles," as depicted in Figures~\ref{fig:Needle-San-Francisco} and~\ref{fig:Needle-New-York}. \ours~demonstrates robust model performance across varying token lengths and depths, maintaining efficacy at both 50\% and 25\% GPU memory budgets. Only a marginal degradation was observed at 12.5\% memory budget. In contrast, DuoAttention variant exhibited a significant performance loss when the memory budget fell below 25\%. These visualizations underscore the robustness of our approach in preserving retrieval capabilities under extreme KV cache compression conditions.

\section{Enhanced prefilling with block-wise attention}  % 附录B：补充图表

Our proposed method, \ours, can be further enhanced by integration with other efficient long-context LLM inference techniques during the prefilling stage.

Inspired by Star Attention, our approach utilizes a block-wise decomposition strategy. This breaks down computationally intensive global attention into more manageable, localized computations. For prefilling, attention is computed exclusively between a designated anchor block and the current processing block. This approximated attention computation significantly boosts computational efficiency. Furthermore, when a block operation concludes, the quantization of the key cache can further enhance efficiency without compromising performance.

\begin{figure}
    \centering
    \includegraphics[width=.9\linewidth]{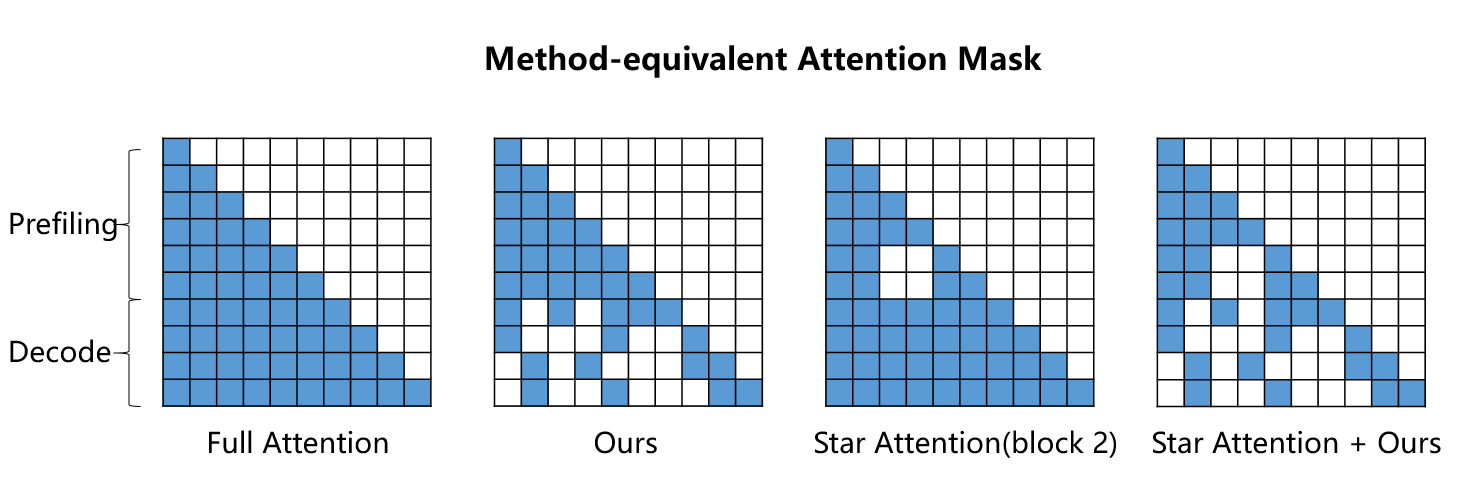}
    \caption{Attention mask patterns illustrating the integration of  \ours~with block-wise attention to achieve efficient prefilling and decoding.}
    \label{fig:Attn_mask}
\end{figure}

As illustrated in Figure \ref{fig:Attn_mask}, the attention mask pattern equivalent to this approach clearly demonstrates how these localized computations effectively approximate global attention behavior. This design allows the model to capture essential long-range dependencies, crucial for understanding complex contexts, even though computations are performed on smaller, manageable blocks. This approximation is vital for deploying large language models on devices with limited memory and computational resources.

\begin{figure}
    \centering
    \includegraphics[width=.9\linewidth]{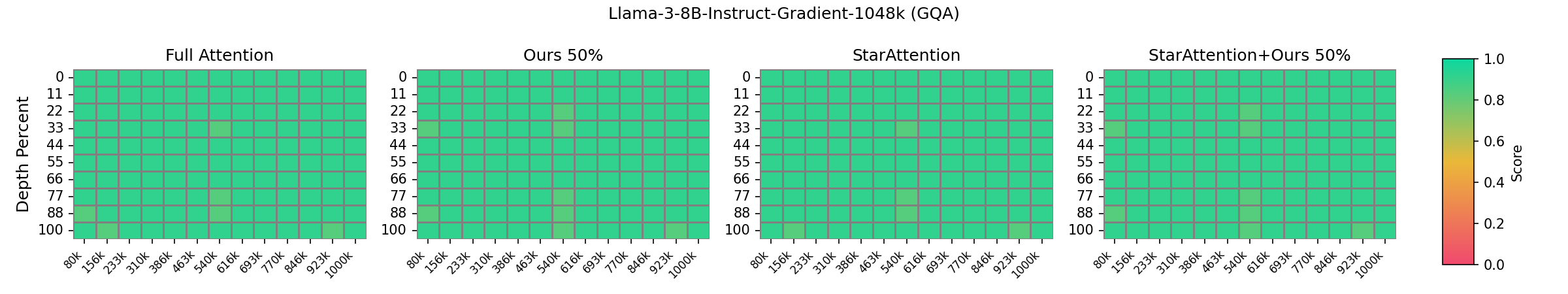}    
    \caption{Visualization of NIAH test of \ours~enhanced with block-wise attention in prefilling stage.}
    \label{fig:starniah}
\end{figure}

Figure~\ref{fig:starniah} demonstrates that by combining \ours~with the sparsification technique during the prefilling phase, our method achieves performance on the NIAH task equivalent to full attention.

\section{Experiment setup details}
\textbf{Hardware Platform} The experiments were conducted on Linux servers running Ubuntu 22.04, configured with: 2 $\times$ Intel Xeon 8358P CPUs, 1TB RAM, and 8 $\times$ NVIDIA A100  GPUs with 80GB memory.

\textbf{Software} Our algorithm was implemented using PyTorch (version 2.5.1+cu121) and HuggingFace Transformers (version 4.45.2).
\end{document}